\pdfoutput=1

\documentclass[11pt]{article}

\usepackage[]{acl}
\usepackage{times}
\usepackage{latexsym}

\usepackage[T1]{fontenc}

\usepackage{pifont}
\newcommand{\cmark}{\ding{51}} 
\newcommand{\xmark}{\ding{55}} 
\usepackage{array,longtable,ragged2e}
\definecolor{myBlueBg}{RGB}{211,223,241}
\definecolor{myBlueTx}{RGB}{86,115,188}
\definecolor{myBlueGreenBg}{RGB}{190,234,228}
\definecolor{myBlueGreenTx}{RGB}{32,133,143}
\definecolor{myApricotBg}{RGB}{245,236,225}
\definecolor{myApricotTx}{RGB}{201,138,128}
\definecolor{myVioletRedBg}{RGB}{234,206,229}
\definecolor{myVioletRedTx}{RGB}{189,64,130}
\definecolor{myYellowGreenBg}{RGB}{229,240,230}
\definecolor{myYellowGreenTx}{RGB}{128,173,135}
\definecolor{myRoyalBlueBg}{RGB}{224,229,233}
\definecolor{myRoyalBlueTx}{RGB}{90,157,181}
\definecolor{myVioletBg}{RGB}{227,219,244}
\definecolor{myVioletTx}{RGB}{102,85,146}
\definecolor{myRoseBg}{RGB}{255,245,221}
\definecolor{myRoseTx}{RGB}{167,126,118}

\usepackage{caption}
\usepackage{subcaption}

\usepackage{microtype}

\usepackage{graphicx}
\usepackage{amsmath}
\usepackage{amsthm}
\usepackage{booktabs}
\usepackage{algorithm}
\usepackage{algorithmic}
\usepackage[switch]{lineno}
\usepackage{seqsplit}
\usepackage[inline]{enumitem}
\usepackage{tabularx}
\usepackage{booktabs}
\usepackage{multirow}
\usepackage{float}

\title{StatBot.Swiss: Bilingual Open Data Exploration in Natural Language}

\author{
Farhad Nooralahzadeh$^{1\dagger}$,
Yi Zhang$^{1\dagger}$,
Ellery Smith${^1}$,
Sabine Maennel${^2}$\\
{\bf Cyril Matthey-Doret}${^2}$,
{\bf Rapha\"el de Fondville}${^3}$,
{\bf Kurt Stockinger}${^1}$\\
$^{^1}$Zurich University of Applied Sciences, Switzerland\\
$^{2}$Swiss Data Science Center, Switzerland\\
$^{3}$Federal Statistical Office, Switzerland\\
 \texttt{\small\{farhad.nooralahzadeh, yi.zhang, kurt.stockinger\}@zhaw.ch}
}

\begin{document}

\maketitle
\renewcommand{\thefootnote}{\fnsymbol{footnote}}
\footnotetext[2]{Equal contribution.}
\renewcommand{\thefootnote}{\arabic{footnote}}
\begin{abstract}
The potential for improvements brought by Large Language Models (LLMs) in Text-to-SQL systems is mostly assessed on monolingual English datasets. However, LLMs' performance for other languages remains vastly unexplored.
In this work, we release the StatBot.Swiss dataset, the \emph{first bilingual benchmark for evaluating Text-to-SQL systems} based on real-world applications. The StatBot.Swiss dataset contains 455 natural language/SQL-pairs over 35 big databases with varying level of complexity for both English and German.
We evaluate the performance of state-of-the-art LLMs such as GPT-3.5-Turbo 
and mixtral-8x7b-instruct for the Text-to-SQL translation task using an in-context learning approach. 
Our experimental analysis illustrates that current LLMs struggle to generalize well in generating SQL queries on our novel bilingual dataset~\footnote{We release our data and code to the community at \url{https://github.com/dscc-admin-ch/statbot.swiss}}.

\end{abstract}

\section{Introduction}
Switzerland is a multilingual country, officially recognizing four national languages: German, French, Italian, and Romansh.  Multilingualism is a pillar of the country's identity, ensuring that all citizens have equal access to public service, education, and any other information, regardless of their linguistic background. Switzerland started offering open government data, and the supply continues to grow, spurred on, among other things, by open-by-default regulations of the federal government, and initiatives from cantons and municipalities. 

While the \emph{opendata.swiss} initiative implements the once-only principle by offering a central catalog for all available Swiss open government data, these datasets are often not standardized across administration levels and neither are methods of collection, compilation, and processing. If data are found, one must then work out and understand the methodological differences in order to know which data are more suitable for the intended usage and more importantly be capable of importing and analyzing the data through statistical software such as a spreadsheet, Python, R or SAS, which all require advanced computing skills.\\
\indent These challenges pose a risk for the democratic processes: the harder it is for a citizen to access an accurate source of information such as national statistics, the more likely a country is to suffer from misinformation.
The StatBot.Swiss project aims to develop a Swiss statistical bot that simplifies access to open government data by allowing interaction with the data directly via natural language.
More precisely, one enters a question and gets
an answer built upon the result of a query on trusted datasets from the opendata.swiss platform.\\
\indent The \emph{main objectives} of this paper are \begin{enumerate*}[label=(\arabic*), itemjoin={{, }}, itemjoin*={{, and }}] 
\item to provide a \emph{real-world bilingual dataset to benchmark Text-to-SQL systems}, namely the StatBot.Swiss dataset, which consists of 455 intricate instances of querying data from 35 large databases, totaling 7.5 GB in size, available in both English and German
\item to \emph{assess the performance of current state-of-the-art pre-trained LLMs} with various prompting strategies and so to establish a strong baseline on the StatBot.Swiss dataset. 
\end{enumerate*}\\
\indent To the best of our knowledge, the StatBot.Swiss dataset is the \emph{first Text-to-SQL benchmark to incorporate English and German languages} within the context of a complex Text-to-SQL benchmark and real-world databases.
We conduct comprehensive evaluations on two pre-trained LLMs, namely GPT-3.5-Turbo-16k,
~\cite{NEURIPS2020_1457c0d6} and Mixtral-8x7B-Instruct-v0.1~\cite{jiang2024mixtral}. Our experimental results show that current models using in-context learning strategies achieve up to 50.58\% execution accuracy and struggle to generate SQL queries giving exact matches on StatBot.Swiss, which shows that 
 multilingual Text-to-SQL systems using state-of-the-art LLMs still lack robustness for reliable applications.

\begin{figure*}[ht]
\centering
\begin{subfigure}{.5\textwidth}
  \centering
  \includegraphics[clip, trim=2cm 2.5cm 0.6cm 2cm, width=1\linewidth]{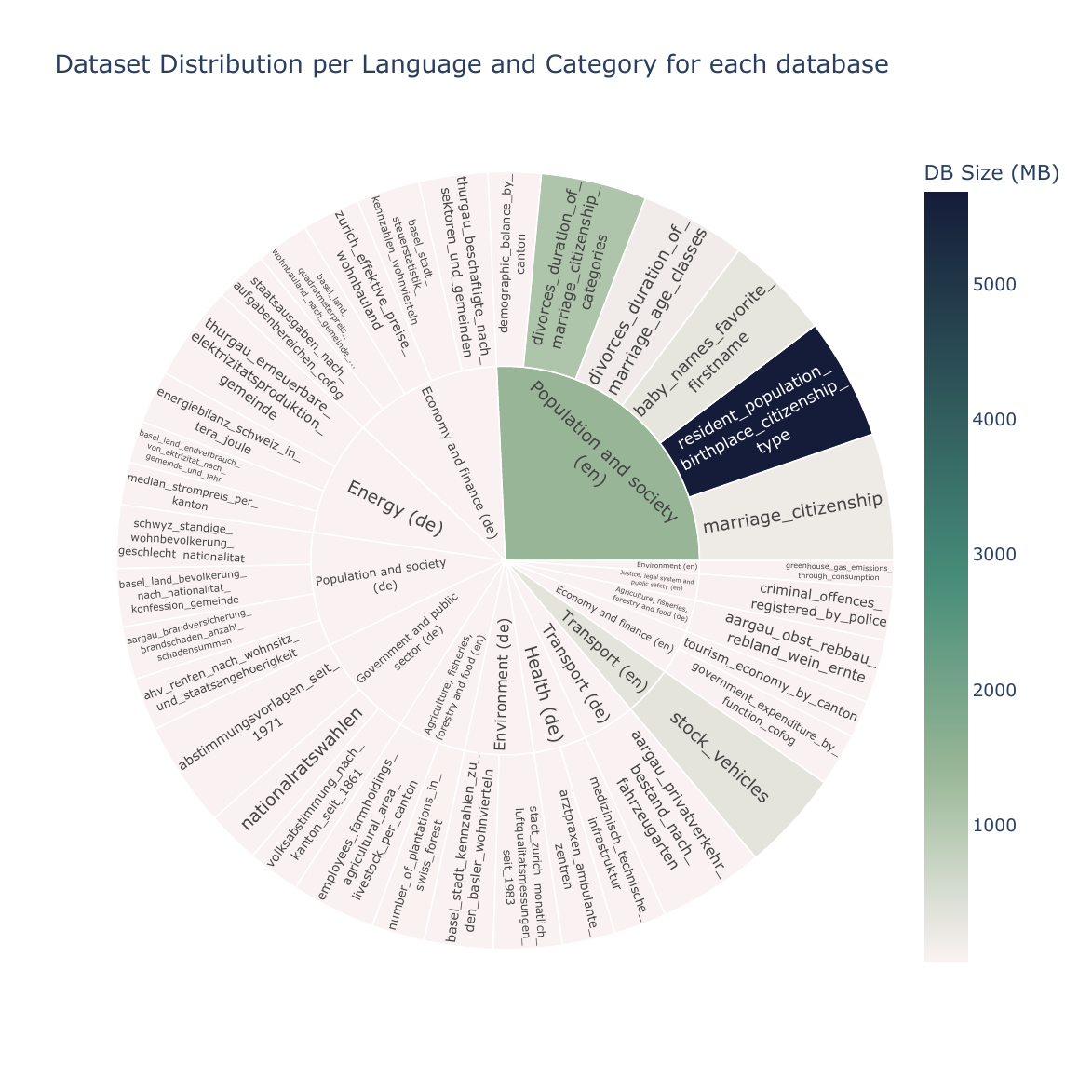}
  \label{fig:data_schema_distribution}
\end{subfigure}%
\begin{subfigure}{.5\textwidth}
  \centering
  \includegraphics[clip, trim=0cm 1.5cm 0cm 1cm, width=1\linewidth]{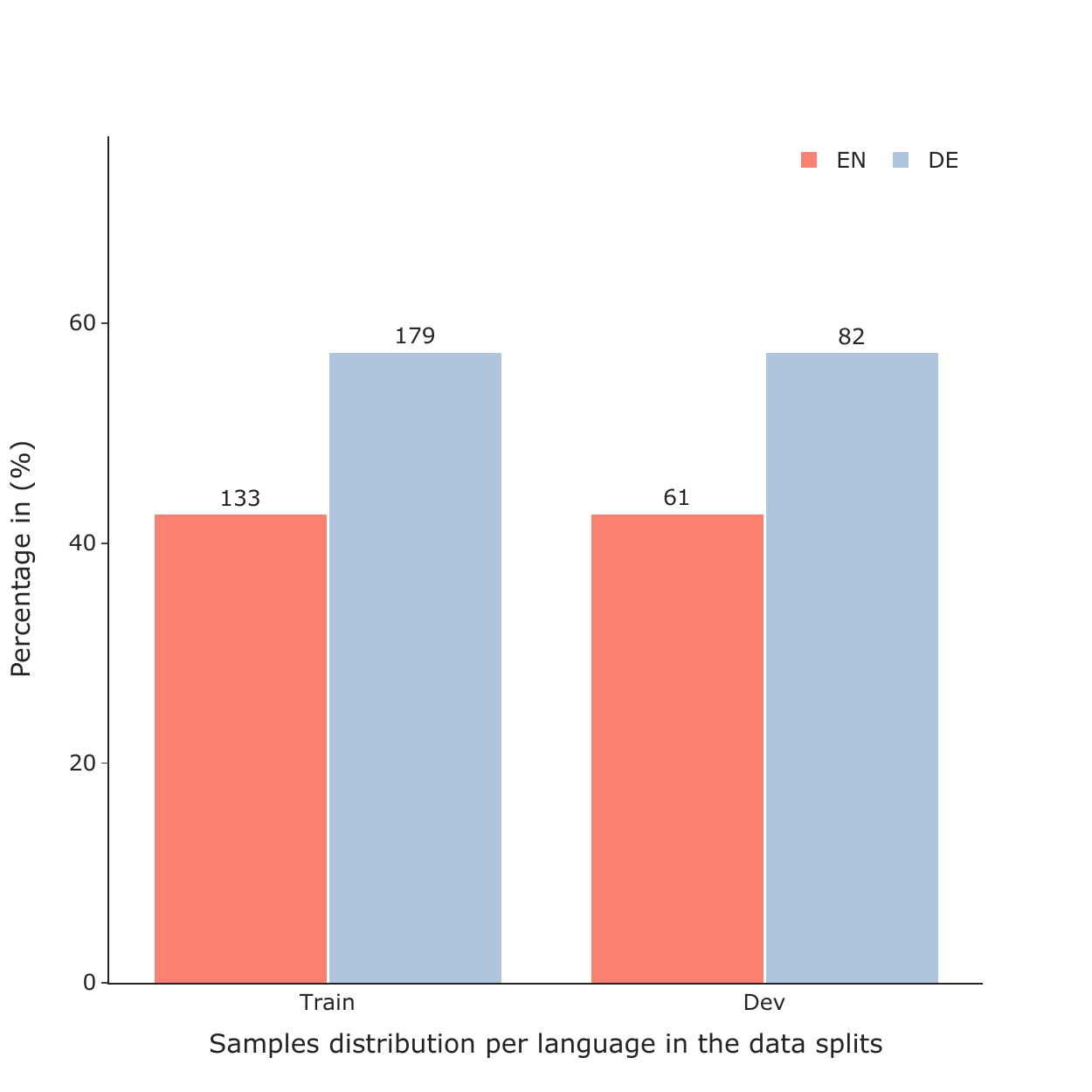}
  \label{fig:lang_per_data_set}
\end{subfigure}
\caption{\textbf{Dataset distribution:} (a) Left: Knowledge domains, (b) Right: Distribution of natural language/SQL-pairs over the train and development sets. EN = English, DE = German. The numbers on top of the bars denote the number of Text-to-SQL pairs.}
\label{fig:dataschema_lang_distribution}
\end{figure*}

\section{Related Work}
\paragraph{Text-to-SQL Dataset} The development of Text-to-SQL datasets and corresponding benchmarks, such as the Spider \cite{yu-etal-2018-spider} and WikiSQL \cite{zhong2017seq2sql} datasets and other domains \cite{deriu2020methodology}, has been instrumental in advancing natural language interfaces to databases. Although much of the initial work focused on the English language, some progress has been achieved in creating resources for other languages. DuSQL \cite{wang-etal-2020-dusql}  brings Text-to-SQL interpretation to Mandarin. PAUQ \cite{bakshandaeva-etal-2022-pauq} fills the gap in the landscape of the Russian language by introducing the complexities associated with Slavic languages to the Spider dataset. ViText2SQL \cite{tuan-nguyen-etal-2020-pilot} is the first public large-scale Text-to-SQL semantic parsing dataset for Vietnamese. 

MultiSpider \cite{10.1609/aaai.v37i11.26499} generalizes the Spider benchmark to multiple languages. However, the drawbacks of MultiSpider are also obvious.  As a translated work from the Spider dataset, the complexity of the database schema and the datasets are strongly limited, and the translation lacks native language expertise. 

Unlike MultiSpider, the StatBot.Swiss benchmark is realistic, complex, and curated by native speakers with proper domain knowledge, while covering both English and German for datasets with trusted sources.

\paragraph{LLMs in Text-to-SQL Translation}
Recently, there has been significant development in employing Large Language Models (LLMs) for the Text-to-SQL task. Several approaches have been suggested to improve the abilities of LLMs by in-context learning techniques \cite{rajkumar2022evaluating,liu2023comprehensive,nan-etal-2023-enhancing} or via semantic hypothesis  re-ranking \cite{von2022improving}. Additionally, methods such as intermediate reasoning steps and self-correction mechanisms have been integrated to enhance the performance of LLMs in Text-to-SQL applications \cite{chen2024teaching,pourreza2023dinsql}. Following these pioneering works, we combine in-context learning to LLMs for the Text-to-SQL task and conduct a comprehensive evaluation of prompt representations over the StatBot.Swiss benchmark.

\section{System Overview}

\subsection{Database preparation} 
\textsc{OpenData.swiss}\footnote{https://opendata.swiss} is the Swiss public administration’s central portal for open government data. As of the date of the submission, the catalog lists more than 10,300 datasets, covering 14 categories, e.g., \textit{health}, \textit{environment}, and \textit{economy}. We meticulously selected 22 German and 13 English datasets from opendata.swiss and subsequently generated corresponding PostgreSQL databases through an automated pipeline. 

\begin{table*}
\setlength\tabcolsep{5pt}
    \centering   
    \resizebox{\textwidth}{!}{
    \begin{tabular}{l|ccccccc}
        \toprule
        \multirow{2}{*}{Dataset}  & \multirow{2}{*}{\#Examples (\#DBs)} & \multirow{2}{*}{\#Tables (\#Rows)/DB} & \multirow{2}{*}{Language} & \multirow{2}{*}{Function} & \multirow{2}{*}{Granularity} & \multirow{2}{*}{Knowledge} & WITH- \\
          &  &  &  &  &  &  & Queries\\
        \midrule
        WikiSQL\cite{zhong2017seq2sql} & 80,654 (26,521)& 1 (17) & EN & \textcolor{red}{\xmark} & \textcolor{red}{\xmark} & \textcolor{red}{\xmark} & \textcolor{red}{\xmark} \\
        SPIDER \cite{yu-etal-2018-spider} & 10,181 (200) & 5.1 (2K) & EN & \textcolor{red}{\xmark} & \textcolor{red}{\xmark} & \textcolor{red}{\xmark} & \textcolor{red}{\xmark} \\
        KaggleDBQA \cite{lee2021kaggledbqa} & 272 (8) & 2.3 (280K) & EN & \textcolor{red}{\xmark} & \textcolor{red}{\xmark} & \textcolor{teal}{\cmark} & \textcolor{red}{\xmark} \\
        ScienceBenchmark \cite{zhang2023sciencebenchmark} & 5,332 (3)& 16.7 (51M) & EN & \textcolor{red}{\xmark} & \textcolor{red}{\xmark} & \textcolor{teal}{\cmark} & \textcolor{red}{\xmark} \\ 
        BIRD \cite{li2023llm} & 12,751 (95) & 7.3 (549K) & EN & \textcolor{teal}{\cmark} & \textcolor{red}{\xmark} & \textcolor{teal}{\cmark} & \textcolor{red}{\xmark} \\ 
        \midrule
        StatBot.Swiss & 455 (35) & 2 (1.4M) & EN, DE & \textcolor{teal}{\cmark} & \textcolor{teal}{\cmark} & \textcolor{teal}{\cmark} & \textcolor{teal}{\cmark} \\
        \bottomrule
    \end{tabular}}
     \caption{Comparison between StatBot.Swiss and existing state-of-the-art Text-to-SQL datasets. \texttt{Function} refers to SQL built-in functions. \texttt{Knowledge} stands for the necessity of external knowledge reasoning from the model. \texttt{Granularity} refers to the degree of specificity or the level of details. \texttt{WITH-Queries} refer to complex sub queries that are broken up into smaller ones. EN = English, DE = German.}
     \label{tab:data_schema_compare}
\end{table*}

\subsection{Data Preparation}
\label{sec:datapreparation}

Each database contains a \emph{fact table} that describes the knowledge domain of the data, and a \emph{dimension table} for the corresponding spatial information (see Database Schema Example in Appendix \ref{app:D}). Note that some statistical information is collected at the municipality level, while other information is collected at the cantonal level (which corresponds to a state). Moreover, the content of the fact tables does not overlap. Finally, one spatial dimensional table is connected to each of the 35 fact tables via a single foreign key constraint.
 
 

\noindent\textbf{Dataset preparation}
For each dataset, our experts analyzed the data sources and formulated natural language questions that the dataset could answer. We then crafted SQL queries to address these questions, executed them, and compared the results with the original dataset. 
Since every table in our database has only a single foreign key - referencing the spatial dimension - the queries that we prepared also target only a single table for each dataset with a join on the spatial unit table. 
 

\noindent\textbf{Dataset statistics}
After manual curation and validation, the dataset includes 455 natural language question (NL)/SQL-pairs covering a total of 35 databases. Figure \ref{fig:dataschema_lang_distribution}(a) presents the distribution of the datasets by languages, spanning a wide range of domains covered by opendata.swiss.
There is, however, an imbalance in the number of NL/SQL-pairs between datasets:
for instance, there are 23 NL/SQL-pairs for the database \texttt{\seqsplit{marriage\_citizenship}}, while only 5 were generated for the database \texttt{\seqsplit{greenhouse\_gas\_emissions\_through\_consumption}}. 
Figure \ref{fig:dataschema_lang_distribution} (b) gives an overview of the sample distribution for the train set and the development set. 

\noindent\textbf{SQL statistics} 
\label{dec:sql_statistics}
In order to evaluate the complexity of the queries, we apply the Spider hardness metric \cite{yu-etal-2018-spider}. However, the StatBot.Swiss dataset includes additional PostgreSQL grammar and syntax features, which are not supported by the Spider hardness evaluator: our dataset includes queries with unencountered levels of complexity in the Spider dataset. For more details see Appendix \ref{app:unknow_spidere_hardness}.
 
Thus, we \emph{extend the Spider-hardness} to categorize the NL/SQL-pairs into five classes, namely the four classes \textit{easy}, \textit{medium}, \textit{hard}, and \textit{extra}, inherited from Spider-hardness with the addition of a fifth category \textit{unknown}. 
In Table \ref{tab:data_schema_compare}, we compare our StatBot.Swiss dataset with other state-of-the-art benchmarks for the Text-to-SQL task using several metrics. Although our new dataset appears simple regarding the number of databases and tables, the training and development datasets (see Section \ref{sec:experimental_setting} for details) are highly complex and cover more realistic natural language questions with external knowledge and more complex SQL syntax than state-of-the-art benchmarks. 
 

\noindent\texttt{Language:} The StatBot.Swiss dataset includes tables in two languages, that is, each table is in English or German. All other datasets are only available in English.

\noindent\texttt{Function:} Only BIRD and our dataset contain built-in functions, e.g., \texttt{CAST()} for type casting or \texttt{ROUND()} for rounding a number. 

\noindent\texttt{Granularity}: It refers to the degree of specificity or the extent to which data are segmented or detailed \citep{1385679,huang2023data}. Data may be depicted at varying levels of granularity, spanning from fine-grained (indicating a high degree of detail, e.g., price of a single item) to coarse-grained (denoting a low level of detail, e.g., subtotal or total price of all items). The choice of granularity depends on the objectives of the analysis and the information required from the data.
As an illustration, the table \texttt{\seqsplit{criminal\_offences\_registered\_by\_police}} contains records at both the coarse and detailed levels querying data at the highest level of detail with \texttt{WHERE \seqsplit{offence\_criminal\_code = 'Offence - total'}} and also including detailed values for each \texttt{\seqsplit{offence\_criminal\_code}}.

\noindent\texttt{Knowledge:} It is defined as the external knowledge reasoning by \citealp{li2023llm}. The reasoning types comprise \textit{Domain knowledge}, \textit{Numeric Computing}, \textit{Synonym} and \textit{Value Illustration}. \citealp{li2023llm} also suggests that a comprehensive understanding of the database content is imperative for addressing more complex real-world queries in Text-to-SQL applications. 

\noindent\texttt{WITH-Queries:} Finally, the StatBot.Swiss records cover very complex query syntax, e.g. \texttt{WITH}-queries allowing users to break down complex queries into several smaller subqueries. 
 
\subsection{Text-to-SQL Translation}
The Text-to-SQL translation task aims to match a natural language question with an SQL query that can effectively extract relevant information from a database. Its formal definition is~\cite{wang2020rat}: Given a natural language question
encoded as a sequence of natural language tokens $\mathcal{Q}=\left\{q_1, \ldots, q_{|\mathcal{Q}|}\right\}$ and a relational database schema
$\mathcal{S}=\langle\mathcal{T}, \mathcal{C}\rangle$ where
$\mathcal{T}=\left\{t_1, \ldots, t_{|\mathcal{T}|}\right\}$ is a series of tables,
and 
$ \mathcal{C}=
\left\{c_1, \ldots, c_{|\mathcal{C}|}\right\}$ denotes their corresponding columns,
a Text-to-SQL system is a function $f(\mathcal{Q}, \mathcal{S})$ that outputs
a correct SQL query $\mathcal{Y}=\left\{y_1, \ldots, y_{|\mathcal{Y}|}\right\}$ as a sequence of tokens.

To develop a Text-to-SQL system, a prevalent approach is to collect labeled data and train a model via supervised learning~\citep{scholak-etal-2021-picard,9458778}. Although effective, this approach requires a considerable amount of training data, which is time- and resource-consuming: annotating SQL queries
requires SQL-specific expertise. 

As an alternative to supervised learning, \emph{in-context learning} (ICL)~\citep{NEURIPS2020_1457c0d6}, an emergent method of large language models (LLMs), alleviates the need for large training datasets: with only a few examples, ICL enables LLMs to demonstrate performance comparable to, if not better, than fully supervised models for many NLP downstream tasks. When applied to the task of Text-to-SQL, ICL achieves encouraging
results~\cite{liu2023comprehensive,nan-etal-2023-enhancing}. 

Following this line of research, we formulate a Text-to-SQL task as $f(\mathcal{Q}, \mathcal{S},\mathcal{E},\mathcal{P})$, where $f$ is a $LLM$, and $\mathcal{E}$ is a set of $m$ in-context exemplars as ${(\mathcal{Q}_i
, \mathcal{Y}i)_{i<m}, \mathcal{Q}_i}\neq\mathcal{Q}\}$. $\mathcal{P}$ is a textual template to represent the overall input, i.e.  $\mathcal{Q}, \mathcal{S},\mathcal{E}$, as a prompt and is fed into the LLM.
We now describe the structure of our prompt design for our Text-to-SQL system.

\noindent\textbf{Database information}
Providing prior knowledge, i.e. examples, about an underlying task can aid the generation process of LLMs. Particularly, in the Text-to-SQL task, inclusion of a database's metadata such as table relationships and variable encoding  is crucial for enabling effective prompting \cite{chang2023how,rajkumar2022evaluating}.

We opt for a \emph{textual representation of the database information}, as suggested by \citet{rajkumar2022evaluating}. In particular, we rely on a CREATE-statement for the initial table creation; see the example prompt in Appendix \ref{app:B} for more details. This representation encompasses specific data type information for each column and integrates all foreign key constraint details within the database. Furthermore, database content is partially included. Specifically, the strategy appends a number $r > 0$ of example rows from each table; see Appendix \ref{app:A} for prompts with $r=5$. 

Furthermore, in line with the findings of \citet{huang2023data}, and \citet{nan-etal-2023-enhancing} who demonstrated the efficacy of schema augmentation through metadata via in-context learning for the Text-to-SQL task, we \emph{enhance the representation of the database structure}. This augmentation involves integrating metadata column information, such as column name and column description, as illustrated in Appendix \ref{app:B}.
\newline
\noindent\textbf{Selection of exemplars $\mathcal{E}$ }\label{db-schema}
 In-context learning enables LLMs to improve their performance on the Text-to-SQL tasks with a small number of training data, with or without a small number of example pairs containing natural language questions and their corresponding SQL representations.
We consider two widely used settings for in-context learning in Text-to-SQL:
(i) \textbf{Zero-shot}: In this context, the evaluation focuses on the Text-to-SQL capability of pretrained LLMs, where the goal is to infer the relationship between a given question and SQL directly from unknown LLM's training dataset. This approach does not leverage any kind of task specific examples; the input prompt is limited to a natural language question along with its corresponding database metadata. The zero-shot setting
is used to directly assess the Text-to-SQL capability of pretrained LLMs \citep{rajkumar2022evaluating,chang2023how}. (ii) \textbf{Few-shot} In this scenario, the LLMs' prompts include examples from a benchmark such as Statbot.swiss. We include a small number of pairs consisting of natural language questions and their corresponding SQL examples, which are inserted between the representation of the database and the target question. The aim is to assess the Text-to-SQL performance of LLMs using a limited number of training data. When employing few-shot learning, a crucial aspect to consider is the careful selection of a subset of demonstrations from a pool of annotated examples for each test instance. This particular design choice plays a significant role and can influence the overall performance of ICL~\citep{liu-etal-2022-makes,nan-etal-2023-enhancing}.

In the few-shot scenario, we established various subset selection methods, including \emph{random} and \emph{similarity-based} strategies. To implement the latter, we initially transformed both the in-domain training questions, i.e., those from the same database as the target question, and the test question into vectors using a sentence encoder. Through cosine similarity we then identified and selected the top $m$ training examples that exhibited the highest similarity to the target question as exemplars. Appendix \ref{app:B} displays a prompt example comprising two (i.e. $m=2$) chosen training instances.

\section{Experiments and Results}
\subsection{Experimental Settings}
\label{sec:experimental_setting}
\paragraph{Dataset} We randomly split the StatBot.Swiss dataset into training and development datasets. Specifically, the dataset is divided into a development dataset constituting 30\% of the total data and a training dataset comprising the remaining 70\%. The splitting is stratified to ensure that the proportion of records with varying degrees of complexity, as introduced in Section \ref{dec:sql_statistics}, is maintained equal for all combinations of partition elements and databases. 


Table \ref{tab:hardness_distribution} in Appendix \ref{app:A} lists the hardness distribution across the multilingual databases for each language in the train, dev, and each individual Statbot.swiss datasets. 

\paragraph{Large Language Models} We examine different strategies for in-context learning using two large language models. To accommodate for extended context lengths in some of our scenarios, i.e., more than 4096 tokens, we restrict our experiment to the GPT-3.5-Turbo-16k model\footnote{Using the API at https://openai.com/api} and the Mixtral-8x7B Instruct model\footnote{https://huggingface.co/mistralai. The model supports up to a 32k context window.}.
To reduce the computational and memory costs of running inferences using the Mixtral model, we apply QLoRA \cite{dettmers2023qlora} with 4-bit quantization and run the inferencing on two \textit{nVidia A100-40GB} GPUs. The LLMs' API temperature is configured to 0, indicating the use of a greedy decoding strategy. 

\paragraph{Evaluation Metrics} \label{sec:eval} We employ the frequently used evaluation metric \emph{execution accuracy} (EA), which computes the percentage of the system's correctly generated SQL statements (predictions), whose execution results correspond to the results of the gold standard SQLs, i.e. the exact output of the query in the benchmark. However, \citealp{Floratou2024} suggest that the original EA may underestimate the overall accuracy due the ambiguity of the queries. Therefore, the evaluation metric should ensure that the generated queries not only produce exact matches,
but also meet the underlying purpose of the user’s
natural language query. 

From a chatbot perspective, the nature of user questions can often encompass a range of acceptable queries. To illustrate this issue with a simple example, let us analyze the question "What canton got the most money from tourism in 2016?" which targets the table \texttt{\seqsplit{tourism\_economy\_by\_canton}}. This question is by nature uncertain, and a user could be equally content with responses that include either [\textit{canton\_name}] only or [\textit{canton\_name, mio\_chf\_gross\_value\_added\_of\_tourism }], which also includes the corresponding funds value. The ground truth from Statbot.swiss for this question outputs only [\textit{canton\_name}]. With EA, a query output with [\textit{canton\_name, \seqsplit{mio\_chf\_gross\_value\_added\_of\_tourism}}] will be considered a \textit{false} answer. 

This complexity makes conventional metrics based on string-matching or execution-matching inadequate.
We start by renaming EA as \emph{strict execution accuracy} ($\text{EA}_{\text{strict}}$) and then introduce  \emph{soft execution accuracy} ($\text{EA}_{\text{soft}}$) and \emph{partial execution accuracy} ($\text{EA}_{\text{partial}}$) for more effective outcome evaluation of certain ambiguous or uncertain questions:

Given $N$ NL/SQL-pairs, we have
$
    \text{EA}_{k} = N^{-1}{\sum_{n=1}^{N}I_{k}(r_n, \hat{r}_n)}
$
where $r_n$, respectively $\hat{r}_n$, are the result set of the ground truth, respectively the system's prediction, $k \in \{\text{strict}, \text{soft}, \text{partial}\}$, and  $I_{k}$ is an indicator function defined as\\ \indent ${I_{\text{strict}}(r_n, \hat{r}_n) = \begin{cases}
        1, & \text{if $r_n = \hat{r}_n$} \\
        0, & \text{otherwise}
    \end{cases}}$
   \indent  ${I_{\text{soft}}(r_n, \hat{r}_n) = \begin{cases}
        1, & \text{if $r_n \subseteq \hat{r}_n$} \\
        0, & \text{otherwise}
    \end{cases}}$  \indent ${I_{\text{partial}}(r_n, \hat{r}_n) = \begin{cases}
        1, & \text{if $r_n \subseteq \hat{r}_n$} \text{ or $\hat{r}_n \subseteq r_n$}\\
        0, & \text{otherwise}
    \end{cases}}$
    
While $\text{EA}_\text{strict}$ might underestimate the fraction of correct answers, its partial and soft versions tend to overestimate the overall system's performance: for the final user, the true performance is likely to lie in-between these two types of metrics.


    


\paragraph{In-context Learning (ICL) Strategies}
We assess the performance of the following ICL strategies for the Text-to-SQL task on the StatBot.Swiss dataset in both zero-shot and few-shot scenarios.

\textbf{Zero-shot} (Baseline): We use the standard prompt for the Text-to-SQL task along with the database information described in Section \ref{db-schema}. The database information is stored as a text representation in the input prompt without any demonstration examples. 

\textbf{Few-shot}: Building upon the zero-shot setting, we select $m \in \{1,2,3,4,5,6,8\}$ NL/SQL-pairs from the Statbot.swiss training set and insert them between the representation of the database and the target question. We pick these instances employing the following approaches:
\begin{enumerate*}[label=(\roman*), itemjoin={{, }}, itemjoin*={{ and }}]
 \item \textbf{Random Selection}: Randomly selecting demonstration examples from the training data, where both the NL-questions and the SQL-queries relate to the same dataset,
 \item \textbf{Similarity-based Selection}: Demonstrative examples are chosen based on their cosine similarity scores as described in Section \ref{db-schema} within the training set, ensuring that both the examples and the target question are collected from the same database. 
 
 Since our dataset comprises both English and German NL/SQL-pairs, we employ the multilingual sentence transformers model \citep{reimers-2020-multilingual-sentence-bert}, specifically the \texttt{\seqsplit{distiluse-base-multilingual-cased-v2}} from HuggingFace model hub  \footnote{https://huggingface.co/sentence-transformers}, to encode natural language questions.
    \end{enumerate*}

\noindent\textbf{Instruction Tuning:}  
 We further fine-tune the Mixtral model using the training dataset and the standard instructions for the Text-to-SQL task, incorporating the database information outlined in Section \ref{db-schema}. Then, we assess the instruction-tuned models in a manner similar to the ICL strategies, including both zero-shot and few-shot evaluations. The model is trained for 200 epochs under Low-Rank Adaptation (LoRA) and 4-bit quantization with a learning rate of $5e^{-5}$. The batch size per device is 1, with gradient accumulation steps set to 16 and epochs set to 100. 
 The fine-tuning is performed on a single NVIDIA A100 40GB GPU for approximately 20 hours.
\subsection{Results}
\begin{figure}[t]
\centering
  \includegraphics[width=1\linewidth]{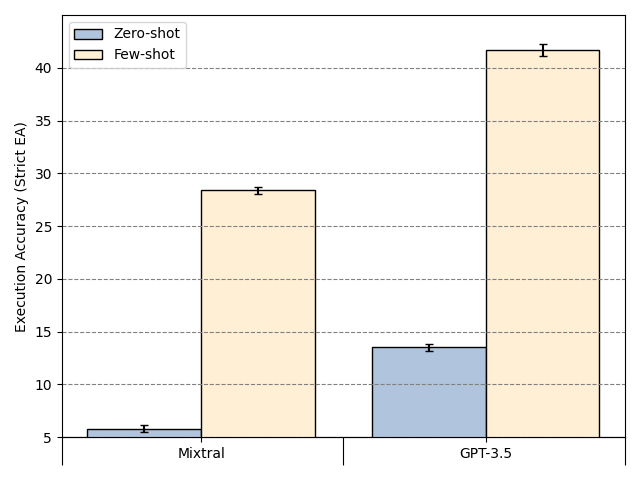}
  \caption{Mean strict execution accuracy : zero-shot and few-shot for GPT-3.5 ($m = 5$) and Mixtral ($m = 6$) models using similarity-based selection where the number of examples are chosen to maximize $\text{EA}_\text{strict}$.}
  \label{fig:result_zero-few}
\end{figure}

\begin{figure*}[hpt]
\centering
\begin{subfigure}{.5\textwidth}
  \centering
  \includegraphics[width=.995\textwidth]{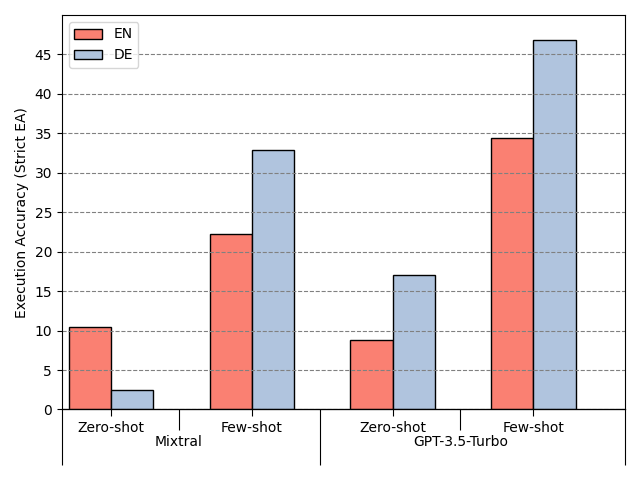}
\end{subfigure}%
\begin{subfigure}{.5\textwidth}
  \centering
  \includegraphics[width=1\textwidth]{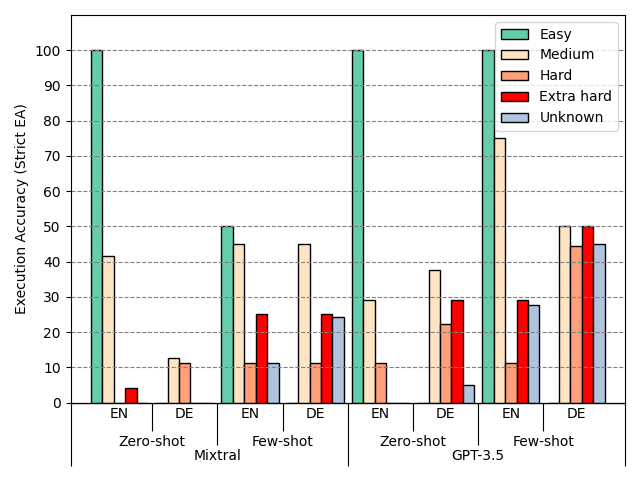}
 
\end{subfigure}

\caption{(Left) Strict execution accuracy ($\text{EA}_\text{strict}$) for each language. (Right) $\text{EA}_\text{strict}$ for each language per query hardness level. All metrics are computed on the development set for zero-shot and few-shot prompting strategies (6-shot in Mixtral, 5-shot in GPT-3.5).}
 \label{fig:result_lang_hardness}
\end{figure*}

In this section, we present a comprehensive analysis of various prompting strategies and their effectiveness across the StatBot.Swiss datasets.  
To make the right conclusions when comparing different scenarios, we calculate the mean and standard deviation (std) of the evaluation metrics using $3$ independent LLMs' queries over the development set with identical prompting strategies. Figures \ref{fig:result_zero-few} and \ref{fig:result_lang_hardness} give a summary of the performance results. We report the detailed estimated evaluation metrics for all models with various demonstration selection strategies and both languages in Tables \ref{tab:table-result-shots}, and \ref{tab:table-result-hardness} of Appendix \ref{app:E}. 

It is worth noting that the instruction-tuned Mixtral model significantly underperformed compared to the ICL strategies evaluation of the original Mixtral model. Consequently, we do not provide further details on the results of this experiment.

\noindent\textbf{Model Performance}: Looking at the \emph{model performance across various shot levels and selection methods}, GPT-3.5 consistently demonstrates superior performance compared to Mixtral, achieving higher mean accuracy scores. Specifically, in terms of $\text{EA}_\text{strict}$, GPT-3.5 achieves a superior result of at most $41.68$\% using $5$ examples, whereas Mixtral is left behind with at most $28.39$\% using $6$ shots. Taking into account the concept of correctness rather than exact matching of the generated SQL query, it can be stated that in the optimal few-shot scenario, GPT-3.5 achieves an $\text{EA}_\text{partial}$ of $50.07$\%.

\noindent\textbf{Zero-shot vs. Few-shot Learning}: In \emph{zero-shot learning}, where no additional labeled examples are provided, both models perform relatively poorly. GPT-3.5 achieves an $\text{EA}_\text{strict}$ of $13.52$\%, outperforming the result by Mixtral at $5.82$\%. \emph{Few-shot learning} significantly improves model performance, demonstrating the effectiveness of providing a small number of labeled examples. Upon providing a small number of labeled examples (one-shot setting), both models show considerable improvement. GPT-3.5 achieves an $\text{EA}_\text{strict}$ of $33.57$\%, resulting in an improvement of approximately $+20.05$\% points compared to the zero-shot setting. Mixtral achieves an $\text{EA}_\text{strict}$ of $16.92$\%, indicating an improvement of approximately $+11$\% compared to the zero-shot setting.

\noindent\textbf{Effect of Selection Method}: The \emph{Similarity selection} method generally outperforms the \emph{Random} selection method, particularly in few-shot learning scenarios. Selecting examples based on their similarity to the natural language questions provides more relevant training examples in ICL for LLMs, leading to better model performance.

\noindent\textbf{Impact of Exemplars Number on Model Performance}:
Transitioning from zero-shot to few-shot learning leads to a significant improvement in model performance across both GPT-3.5 and Mixtral models. On average, each additional shot results in approximately a $5$-$10$\% increase in mean $\text{EA}_\text{strict}$, underscoring the importance of providing labeled examples during the inference. However, this improvement is followed by a significant decrease once the number of examples reaches a certain threshold: maximal performance is achieved with $5$ examples in GPT-3.5, and $6$ examples in Mixtral using similarity-based selection.

\noindent\textbf{Language and Hardness Level}: 
When differentiating by language, with zero-shot learning, GPT-3.5 achieves an $\text{EA}_\text{strict}$ of $8.75$\% for English (EN), whereas Mixtral shows a slightly higher accuracy at $10.39$\%. For German (DE), GPT-3.5 records an $\text{EA}_\text{strict}$ of $17.07$\%, compared to Mixtral's $2.44$\%.
 
Transitioning to the few-shot setting, GPT-3.5 demonstrates significant superiority over Mixtral for English (EN), achieving an $\text{EA}_\text{strict}$ of $34.43$\%, as opposed to Mixtral's $22.29$\%. Similarly, in German (DE), GPT-3.5 exhibits better performance with an $\text{EA}_\text{strict}$ of $46.83$\%, contrasting Mixtral's $32.93$\%.

Moreover, we analyze the Type-Token Ratios (TTR) and the token length of both English and German questions respectively, revealing that the German dataset exhibits greater linguistic diversity and consequently appears more challenging. This finding is inspired by prior research on the difficulty and discrimination of natural language questions by \cite{byrd-srivastava-2022-predicting}, and on cross-lingual summarization by \cite{wang-etal-2023-understanding}, both of which assess lexical richness and diversity. TTR, defined as the ratio of unique tokens to the total number of tokens, serves as a measure of linguistic diversity. The average TTR for German questions is $0.961$, compared to $0.927$ for English questions. Furthermore, the average token length for German questions is $15.6$, whereas for English questions, it is $15.13$. These indices unveil that German questions in the dataset are more diverse and, therefore, potentially more difficult.

The results in Figure \ref{fig:result_lang_hardness} also show that both models exhibit higher performance in German than in English,
despite the German dataset being inherently more difficult. This trend persists even when the difficulty level of SQL within the development dataset is taken into account. This can be attributed to the fact that the German questions were written by native speakers, while the English questions were not, and thus suggesting potential avenues for future research in paraphrasing and linguistic characteristics of both datasets.

Furthermore, Figure \ref{fig:result_lang_hardness} (right) and Table \ref{tab:table-result-hardness} show that GPT-3.5 is unable to answer English questions classified as extra hard or unknown in the zero-shot setting. Conversely, Mixtral succeeds in providing answers for the English questions, even when their difficulty is categorized as medium or extra hard. However, Mixtral also struggles to answer questions in the hard level category, indicating limitations in handling challenging queries despite some capability in the zero-shot scenario. 

It is apparent that in the Mixtral setting with an easy difficulty level, there is a notable decline in performance when transitioning from zero-shot to few-shot scenarios. This decline stems from the scarcity of data points within our dataset under easy difficulty level. Specifically, there are only two instances present in the development set, both in English. Consequently, even a single failed prediction significantly impacts the local distribution of percentage scores, causing the performance on easy samples to drop from 100\% to 50\%. Upon examining the failed prediction, we identified the root cause as a key term swap between COUNT and DISTINCT during the few-shot ICL execution on Mixtral as illustrated in Appendix \ref{app:C}.

On average, Mixtral demonstrates better performance than GPT-3.5 in the zero-shot scenario for English questions. Nevertheless, the situation is reversed in the German dataset.

In the few-shot scenario, considering the optimal prompting strategies, i.e., 5-shot with GPT-3.5 and 6-shot with Mixtral, both models exhibit the capability to provide exact matches in every category. Nevertheless, it is important to note that although both models demonstrate some proficiency in answering questions within the few-shot setup, they encounter challenges on many occasions, especially when dealing with queries of higher difficulty levels such as hard and extra hard.

\section{Error Analysis}
To delve deeper into the difficulties faced by state-of-the-art LLMs in the Text-to-SQL task using the StatBot.Swiss dataset, we have a closer look at the results of the best performing model, namely GPT-3.5, to perform an error analysis.
We observe that \begin{enumerate*}[label=(\roman*), itemjoin={{, }}, itemjoin*={{ and }}]
    \item the model struggles significantly with inferring the queries involving \texttt{GROUP BY} multi-columns and \texttt{mixed numeric operators}. This leads to a low accuracy of $27.27$\% and $38.89$\% for these queries, respectively. 
    
    These difficulties highlight the need for domain knowledge, which might not always be explicitly represented in the database schema
    \item queries requiring proper built-in \texttt{Function}s or other general SQL keywords are answered slightly better by the LLM, which counts for $40$\% and $44.44$\% accuracy, respectively
    \item the language model results in an accuracy of $50$\% for \texttt{NULL}-values and $66.67$\% for nested \texttt{SELECT}-alias, suggesting that the model has learned well from value illustration and grasped the use of sub-queries; see Appendix \ref{app:B} for examples,
    \item finally, the outcome for \texttt{WITH}-queries and \texttt{SET-operation} queries show an accuracy of $100$\%. However, the statistical significance of this performance is not guaranteed because of the small number of examples available in the dataset.
    For a more detailed understanding, we list NL/SQL-pairs for each type of complex query in Table \ref{tab:complex_query_unknown} of Appendix \ref{app:B}.
 \end{enumerate*}

We particularly investigated datasets in which the underlying LLM fails to answer any of their questions; see Figure \ref{fig:HEA_dataset} in Appendix \ref{app:E}. In the \texttt{\seqsplit{tourism\_economy\_by\_canton}} English dataset, there exists a few false predictions due to the wrong understanding of data granularity. However, others are correctly labeled in terms of usefulness metrics, resulting in $66.67$\% of both $\text{EA}_\text{soft}$ and $\text{EA}_\text{partial}$. Given that each dataset was manually curated by various experts, we found that the observed disparity between $\text{EA}_\text{strict}$ and user's intent-based EA, namely soft and partial, is concentrated in certain datasets, which is in line with the previous assumption that semantic ambiguity is often subjective and individual. This insight of finding also demonstrates the importance of $\text{EA}_\text{soft}$ and $\text{EA}_\text{partial}$ from the experimental perspective.

In the datasets of \texttt{\seqsplit{staatsausgaben\_nach\_aufgabenbereichen\_cofog}} and \texttt{\seqsplit{government\_expenditure\_by\_function\_cofog}}, the knowledge misunderstanding related to the granularity of the data led to the majority of wrongly generated queries.
Upon examining the similarity scores between the target question and the few-shot examples within each dataset, it becomes clear that there is a considerable level of disparity in the natural language questions found within these two datasets.

\section{Conclusions}
Our work highlights the potential and current limitations of Large Language Models (LLMs) using our novel bilingual Text-to-SQL benchmark StatBot.Swiss. While LLMs like GPT-3.5-Turbo and mixtral-8x7b show promising results for relatively simple queries, they face challenges with complex queries, domain knowledge inference, and handling \texttt{NULL} values. The analysis underscores the necessity for enhanced in-context learning in LLMs to incorporate a broader range of Text-to-SQL and database knowledge. The observed disparities in error analysis, especially in datasets with semantic ambiguities and data granularity issues, call for a more nuanced approach to evaluating the performance of Text-to-SQL systems, considering correctness from the user's perspective as apposed to exact matching. 

Our work sets a benchmark for future developments in bilingual Text-to-SQL systems and emphasizes the importance of diverse and complex datasets in advancing LLM capabilities. As the field progresses, refining these models to better understand and execute multi-column grouping, numeric operations, and domain-specific queries will be crucial for realizing their full potential in real-world applications. 

In future work, we aim to include other languages, e.g., French and Italian, in our benchmark and extend towards cross-lingual Text-to-SQL tasks. 

\section{Limitations} 
One major constraint of our study lies in the intrinsic challenges of Text-to-SQL evaluations across bilingual datasets. This includes a limited number of Large Language Models (LLMs) assessed in our experiments, the finite languages chosen, and the complexity and diversity of datasets.

Despite our comprehensive analysis, the limited dataset size, the constrained domain scope, the imbalanced query types, and the predefined hardness may not fully represent the breadth of other real-world applications.

Additionally, there is a lack of natural language/SQL-pairs across languages. That means that our benchmark dataset does not use English questions to query a database in German, or the other way around. 

Lastly, our study assumes the target datasets are known beforehand, which could potentially bias the interpretation and application of the Text-to-SQL models' performance. This presupposition that the user or upstreaming network ideally grasps domain-specific knowledge may not always hold true, especially in diverse real-world scenarios where domain expertise varies widely. Such assumptions are apt to yield experimental outcomes that surpass the actual accuracy in realistic applications.


\section*{Acknowledgments}
This work is the output of the INODE4StatBot. swiss project\footnote{https://www.zhaw.ch/en/research/research-database/project-detailview/projektid/5959/}, funded by the Swiss Federal Statistical Office. We want to thank Patrick Arnecke from the Cantonal Statistical Office Zurich as well as Yara Abu Awad, Christine Chorat and Bertrand Loison from the Swiss Federal Statistical Office for valuable feedback during the project.

\bibliographystyle{acl_natbib}
\bibliography{custom}

\newpage

\appendix
\onecolumn

\section{Database Schema Example}\label{app:D}
\begin{figure}[hbpt]
    \centering
    \includegraphics[clip, trim=13cm 0cm 13cm 0cm, width=0.8\linewidth]{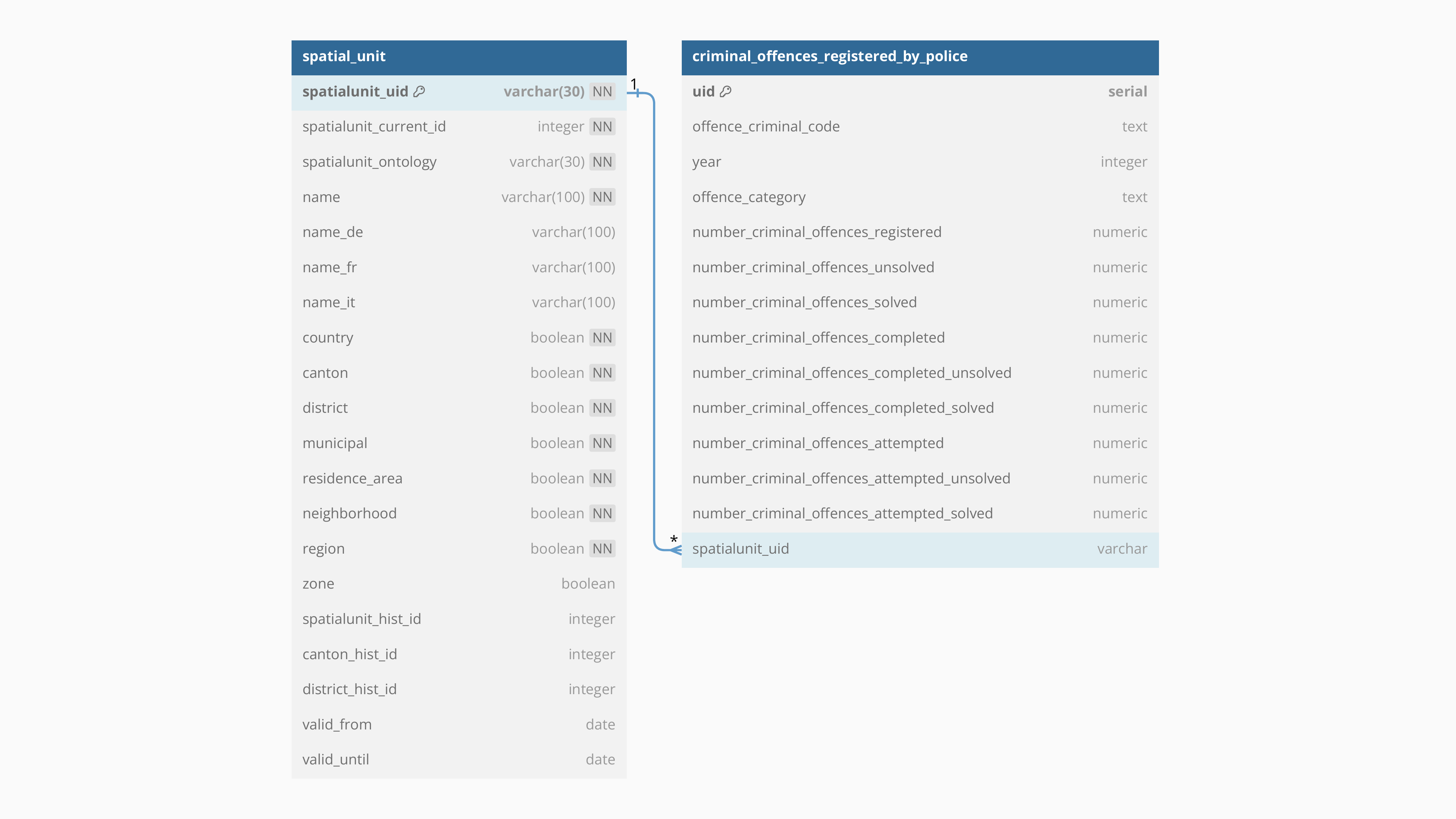}
    \caption{Entity-relationship diagram of the knowledge domain criminal offences. \texttt{spatial\_unit} is the \emph{dimension} table and \texttt{criminal\_offenses\_registered\_by\_police} is the \emph{fact} table. EN = English, \texttt{NN} stands for \texttt{NOT NULL} constraint. Note that the dimension table \texttt{spatial\_unit} contains information about different levels of granularity and thus enables aggregating facts by, e.g. \texttt{municipality}, \texttt{canton} and \texttt{country}. However, note that not all facts contain information about all levels of granularity. For instance, some facts are only collected at municipality level while others are collected a cantonal level.}
    \label{fig:ERD_crimical_offence}
\end{figure}

\begin{figure}[ht!]
    \centering
    \includegraphics[clip, trim=13cm 0cm 13cm 0cm, width=0.8\linewidth]{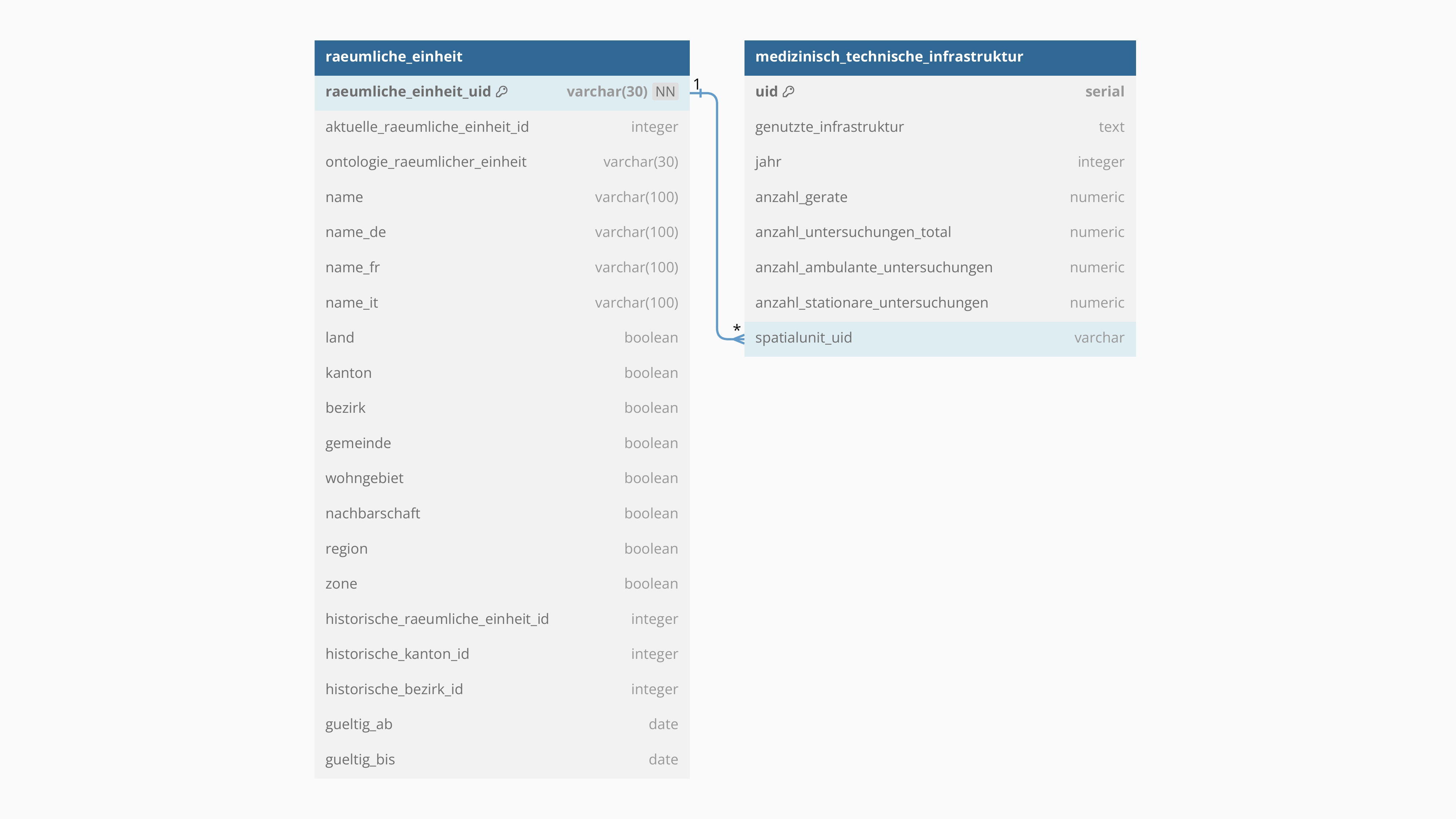}
    \caption{Entity-relationship diagram of the knowledge domain \texttt{medizinisch\_technische\_infrastruktur} {[}DE{]} (in Eng. \texttt{medical technical infrastructure}), where \texttt{NN} stands for \texttt{NOT NULL} constraint. Note that the dimension table \texttt{raeumliche\_einheit} contains information about different levels of granularity and thus enables aggregating facts by, e.g. \texttt{Gemeinde} (in English: municipality), \texttt{Kanton} (in English: canton) and \texttt{Land} (in English: country). However, note that not all facts contain information about all levels of granularity. For instance, some facts are only collected at municipality level while others are collected at cantonal level.}
    \label{fig:ERD_medizinisch_technische}
\end{figure}

\newpage

\section{Queries that Cannot be Analyzed by the Spider Hardness Evaluator}
\label{app:unknow_spidere_hardness}

Here we show the complex query categories of our new dataset that use additional PostgresSQL features that cannot be evaluated by the Spider hardness evaluator:

\begin{enumerate}
    \item \texttt{GROUP BY} or \texttt{ORDER BY} of more than one column, for instance, \texttt{GROUP BY offence\_criminal\_code, \seqsplit{number\_criminal\_offences\_registered}}.
    \item Nested \texttt{SELECT}-query alias, e.g., \texttt{SELECT a.col\_1 FROM (SELECT$\dots$) AS A}. Unlike the column alias, these nested \texttt{SELECT}-queries are often required to break down complex queries into more managable sub queries and may cause errors when omitted.
    \item \texttt{WITH}-queries also enable breaking down larger queries into smaller sub queries.
    \item Built-in \texttt{Function}s except basic \texttt{aggregators}\footnote{The basic \texttt{aggregators} denotes \texttt{COUNT}, \texttt{SUM}, \texttt{MAX}, \texttt{MIN}, and \texttt{AVG}}, e.g., \texttt{CAST}-function for type casting. 
    
    \item Numeric operators mixed with \texttt{aggregator}s, such as \texttt{\seqsplit{100 * SUM(number\_criminal\_offences\_solved) / SUM(number\_criminal\_offences\_registered)}}.
    \item \texttt{SET}-operators to combine sub queries, e.g. \texttt{UNION ALL}, \texttt{INTERSECT ALL}, or \texttt{EXCEPT ALL}.
    \item Special keywords, e.g., \texttt{IN}, \texttt{CASE}.
    \item \texttt{NULL} values.
\end{enumerate}


\section{Query Hardness Distribution of the Bilingual Dataset} \label{app:A}

\begin{table*}[hbp]
\setlength\tabcolsep{2pt}
    \centering
    
    \resizebox{\textwidth}{!}{
    \begin{tabular}{l|ccccc|ccccc}
        \toprule
        & \multicolumn{5}{c|}{English} & \multicolumn{5}{c}{German} \\
         \cmidrule(lr){2-6} \cmidrule(lr){7-11}
          & easy & medium & hard & extra & unknown & easy & medium & hard & extra & unknown \\
         \midrule
         Train & 3 (2.26) & 5 (3.76) & 6 (4.51) & 37 (27.82) & 82 (61.65) & 0 (0.00) & 4 (2.23) & 17 (9.50) & 37 (20.67) & 121 (67.60) \\
         Dev & 2 (3.28) & 8 (13.11) & 9 (14.75) & 24 (39.34) & 18 (29.51) & 0 (0.00) & 8 (9.76) & 9 (10.98) & 24 (29.27) & 41 (50.00)\\
         All & 5 (2.58) & 13 (6.70) & 15 (7.73) & 61 (31.44) & 100 (51.55) & 0 (0.00) & 12 (4.60) & 26 (9.96) & 61 (23.37) & 162 (62.07) \\   
    \end{tabular}}
    \caption{Hardness distribution per language for each dataset. The values in parentheses represent the distribution of samples corresponding to the hardness level across languages and datasets in percentage.}
    \label{tab:hardness_distribution}
\end{table*}

\newpage

\section{Example Prompt} \label{app:B}

Below we show the prompt for two examples (2-shot). First, we show the CREATE TABLE statement (database schema) followed by several example rows (data values) per table. Afterward, we illustrate natural language questions and their corresponding SQL queries. The examples are about the usage of electric cars in certain areas.

\begin{figure*}[hp]
    \centering
    \includegraphics[width=0.75\textwidth]{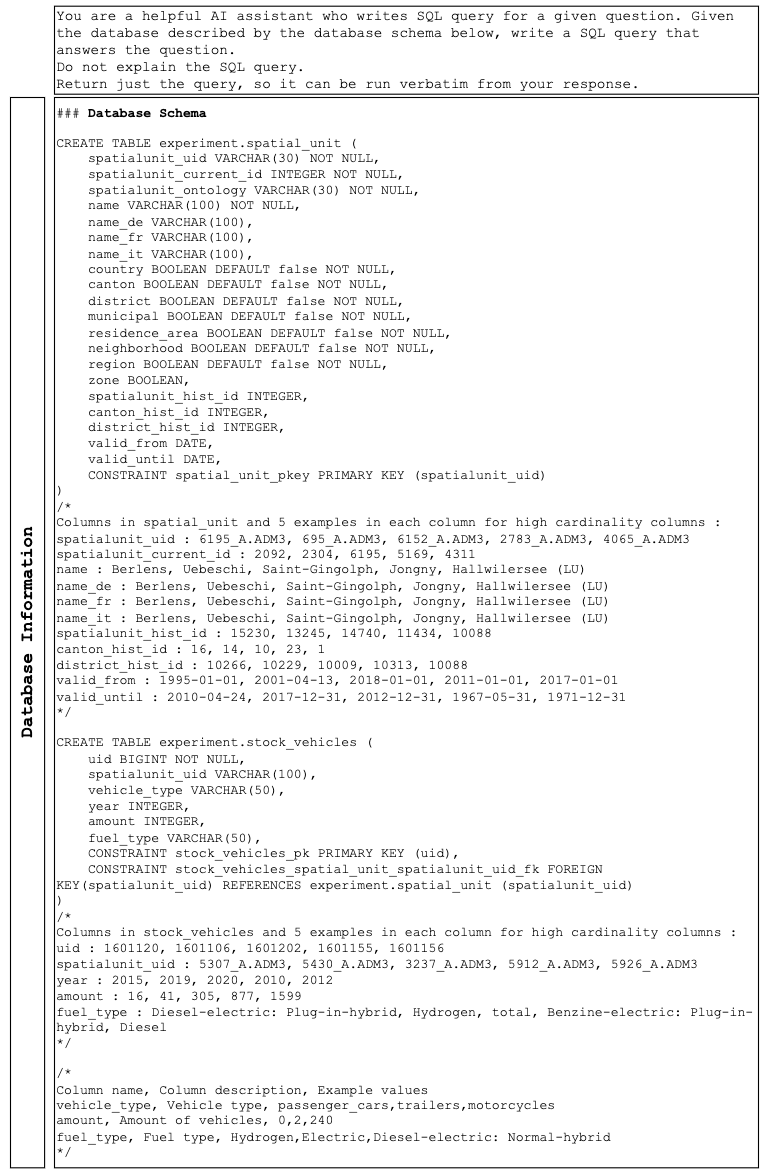}
    \label{fig:prompt-1}
\end{figure*}

\begin{figure*}[thbp]
    \centering
    \includegraphics[width=0.75\textwidth]{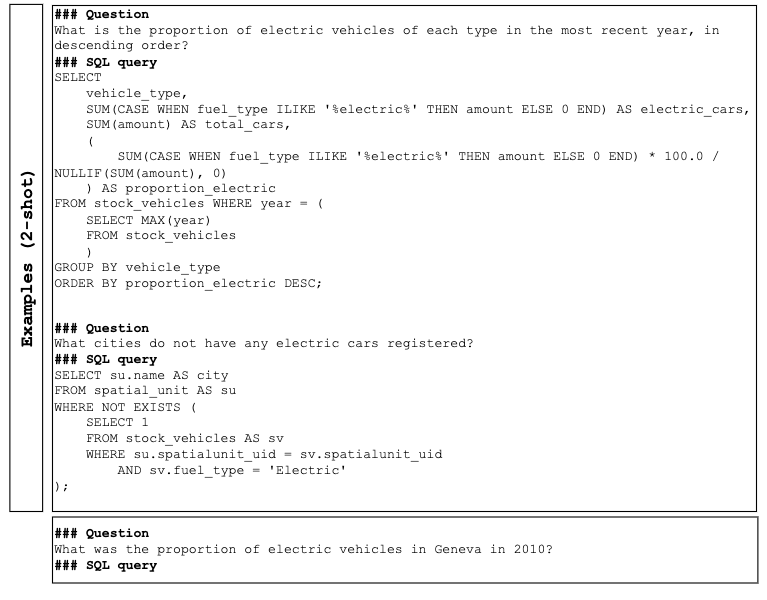}
    \label{fig:prompt-2}
\end{figure*}
\newpage





{\small
\begin{longtable}[hb]{@{} >{\RaggedRight}p{4cm}>{\RaggedRight}p{12cm} @{}}

\caption{Complex SQL query examples with hardness "unknown". Note that these kind of queries cannot be handled by the Spider hardness evaluator due to their high complexity especially in terms of SQL features used.}\label{tab:complex_query_unknown}\\
\toprule
\textbf{Query Types} & \textbf{ {[}db\_id{]} | \textit{Question} | \texttt{Query}}\\
\midrule
\endfirsthead

\multicolumn{2}{@{}l@{}}{Table \ref{tab:complex_query_unknown} (Continued): Complex SQL query examples of hardness "unknown".} \\
\addlinespace
\toprule
\textbf{Query Types}  & \textbf{ {[}db\_id{]} | \textit{Question} | \texttt{Query}}\\
\midrule
\endhead 

\bottomrule
\multicolumn{2}{@{}l@{}}{Table \ref{tab:complex_query_unknown} (End): Complex SQL query examples of hardness "unknown".} \\
\endlastfoot
\colorbox{myBlueBg}{\textcolor{myBlueTx}{1. \texttt{GROUP BY} > 1 column}} & \begin{tabular}[c]{@{}l@{}}{[}volksabstimmung\_nach\_kanton\_seit\_1861{]}\end{tabular} \\
\addlinespace
& \begin{tabular}[c]{@{}p{12cm}@{}}\textit{Welche Kantone haben 2023 gegen das Bundesgesetz über Klimaschutz gestimmt und wieviel Prozent Ja Stimmen gab es dort jeweils?
} \end{tabular} \\
\addlinespace
& \texttt{SELECT S.name\_de AS kanton\_gegen\_klimaschutzgesetz,} \\
& \texttt{T.ja\_in\_prozent}\\ 
& \texttt{FROM volksabstimmung\_nach\_kanton\_seit\_1861 AS T}\\ 
& \texttt{JOIN spatial\_unit AS S} \\ 
& \texttt{ON T.spatialunit\_uid = S.spatialunit\_uid}\\ 
& \texttt{WHERE S.canton = 'TRUE'}\\ 
& \texttt{AND LOWER(T.vorlage) LIKE '\%bundesgesetz\%klimaschutz\%'} \\ 
& \texttt{AND T.jahr = 2023}\\ 
& \texttt{AND T.ja\_in\_prozent \textless{}= 50}\\ 
& \texttt{\textbf{\colorbox{myBlueBg}{\textcolor{myBlueTx}{GROUP BY S.name\_de, T.vorlage, T.jahr, T.ja\_in\_prozent}}};}\\
\midrule
\colorbox{myBlueGreenBg}{\textcolor{myBlueGreenTx}{2. Nested \texttt{SELECT}-Query}} & \begin{tabular}[c]{@{}l@{}}{[}marriage\_citizenship{]}\end{tabular} \\
\colorbox{myBlueGreenBg}{\textcolor{myBlueGreenTx}{  Alias}} \\
\addlinespace
& \begin{tabular}[c]{@{}p{12cm}@{}}\textit{Show me the lowest number of marriages that occurred at the canton level in 1990, where both the wife and husband were from different nationalities?} \end{tabular} \\
\addlinespace
& \texttt{SELECT A.* }\\ 
& \texttt{FROM (SELECT T2.name, T1.citizenship\_category\_husband,} \\
& \texttt{T1.citizenship\_category\_wife, T1.amount}\\ 
& \texttt{FROM marriage\_citizenship AS T1}\\ 
& \texttt{JOIN spatial\_unit AS T2} \\
& \texttt{ON T1.spatialunit\_uid = T2.spatialunit\_uid}\\
& \texttt{WHERE T2.canton = 'True'}\\
& \texttt{AND T1.year = 1990}\\
& \texttt{AND T1.citizenship\_category\_husband = 'Foreign country'}\\ 
& \texttt{AND T1.citizenship\_category\_wife = 'Switzerland'}\\ 
& \texttt{UNION}\\ 
& \texttt{SELECT T2.name, T1.citizenship\_category\_husband, T1.citizenship\_category\_wife, T1.amount}\\ 
& \texttt{FROM marriage\_citizenship as T1}\\ 
& \texttt{JOIN spatial\_unit AS T2} \\ 
& \texttt{ON T1.spatialunit\_uid = T2.spatialunit\_uid}\\
& \texttt{WHERE T2.canton = 'True'}\\ 
& \texttt{AND T1.year = 1990}\\ 
& \texttt{AND T1.citizenship\_category\_husband = 'Switzerland'}\\
& \texttt{AND T1.citizenship\_category\_wife = 'Foreign country') \textbf{\colorbox{myBlueGreenBg}{\textcolor{myBlueGreenTx}{AS A}}} }\\ 
& \texttt{ORDER BY A.amount ASC}\\ 
& \texttt{LIMIT 1;} \\
\midrule
\colorbox{myApricotBg}{\textcolor{myApricotTx}{3. Built-in \texttt{Function}}} & \begin{tabular}[c]{@{}l@{}}{[}basel\_land\_bevolkerung\_nach\_nationalitat\_konfession\_gemeinde{]}\end{tabular} \\
\addlinespace
& \begin{tabular}[c]{@{}p{12cm}@{}}\textit{Welcher Anteil der Bevölkerung von Basel-Landschaft gehörte 2021 einer bekannten Religion an?} \end{tabular} \\
\addlinespace
& \texttt{SELECT 1 - (}\\ 
& \texttt{SUM(\colorbox{myApricotBg}{\textcolor{myApricotTx}{\textbf{CAST(T.anzahl\_unbekannt\_konfession AS FLOAT)}}})} \\ 
& \texttt{/ SUM(T.gesamt\_anzahl\_personen))} \\ 
& \texttt{AS proportion\_known\_religion\_basel\_land\_2021}\\ 
& \texttt{FROM} \\ 
& \texttt{basel\_land\_bevolkerung\_nach\_nationalitat\_konfession\_gemeinde}\\ 
& \texttt{AS T}\\ 
& \texttt{JOIN spatial\_unit AS S ON T.spatialunit\_uid = S.spatialunit\_uid}\\ 
& \texttt{WHERE S.municipal = 'TRUE'} \\     
& \texttt{AND T.jahr = 2021}\\ 
& \texttt{GROUP BY T.jahr;} \\

\midrule

4. \colorbox{myVioletRedBg}{\textcolor{myVioletRedTx}{\texttt{WITH}-Queries}} & \begin{tabular}[c]{@{}l@{}}{[}medizinisch\_technische \_infrastruktur{]}\end{tabular} \\
\addlinespace
& \begin{tabular}[c]{@{}p{12cm}@{}}\textit{Welchem drei Kantone hatte den grössten Zuwachs und Untersuchungen mit medizinischen Geräten zwischen 2013 and 2021 und wieviel hoch war der Zuwachs verteilt auf ambulante und stationäre Untersuchungen und Geräte?} \end{tabular} \\
\addlinespace
& \texttt{\textbf{\colorbox{myVioletRedBg}{\textcolor{myVioletRedTx}{WITH}}} Untersuchungen2013 AS (}\\ 
& \texttt{SELECT SUM(T1.anzahl\_untersuchungen\_total)} \\ 
& \texttt{AS anzahl\_untersuchungen\_gesamt\_2013,}\\  
& \texttt{SUM(T1.anzahl\_ambulante\_untersuchungen)} \\ 
& \texttt{AS anzahl\_ambulante\_untersuchungen\_2013,}\\ 
& \texttt{SUM(T1.anzahl\_stationare\_untersuchungen)} \\ 
& \texttt{AS anzahl\_stationare\_untersuchungen\_2013,}\\   
& \texttt{SUM(T1.anzahl\_gerate) AS anzahl\_gerate\_2013,}\\ 
& \texttt{S1.name\_de AS kanton} \\ 
& \texttt{FROM medizinisch\_technische\_infrastruktur AS T1}\\
& \texttt{JOIN spatial\_unit AS S1} \\ 
& \texttt{ON T1.spatialunit\_uid = S1.spatialunit\_uid} \\
& \texttt{WHERE S1.canton = 'TRUE' AND T1.jahr = 2013} \\
& \texttt{GROUP BY S1.name\_de),}\\ 
& \texttt{Untersuchungen2021 AS (}\\ 
& \texttt{SELECT SUM(T2.anzahl\_untersuchungen\_total)}\\ 
& \texttt{AS anzahl\_untersuchungen\_gesamt\_2021,}\\
& \texttt{SUM(T2.anzahl\_ambulante\_untersuchungen)} \\
& \texttt{AS  anzahl\_ambulante\_untersuchungen\_2021,}\\
& \texttt{SUM(T2.anzahl\_stationare\_untersuchungen)} \\
& \texttt{AS anzahl\_stationare\_untersuchungen\_2021,}\\ 
& \texttt{SUM(T2.anzahl\_gerate) AS anzahl\_gerate\_2021,}\\
& \texttt{S2.name\_de AS kanton2021}\\ 
& \texttt{FROM medizinisch\_technische\_infrastruktur AS T2}\\
& \texttt{JOIN spatial\_unit AS S2} \\
& \texttt{ON T2.spatialunit\_uid = S2.spatialunit\_uid}\\ 
& \texttt{WHERE S2.canton = 'TRUE'}\\ 
& \texttt{AND T2.jahr = 2021}\\ 
& \texttt{GROUP BY S2.name\_de)}\\ 
& \texttt{SELECT U2013.kanton,}\\
& \texttt{U2021.anzahl\_untersuchungen\_gesamt\_2021 - U2013.anzahl\_untersuchungen\_gesamt\_2013} \\ 
& \texttt{AS zuwachs\_untersuchungen\_gesamt,}\\    
& \texttt{U2021.anzahl\_ambulante\_untersuchungen\_2021 - U2013.anzahl\_ambulante\_untersuchungen\_2013} \\ 
& \texttt{AS zuwachs\_ambulante\_untersuchungen,}\\     
& \texttt{U2021.anzahl\_stationare\_untersuchungen\_2021 - U2013.anzahl\_stationare\_untersuchungen\_2013} \\ 
& \texttt{AS zuwachs\_statinonare\_untersuchungen,}\\ 
& \texttt{U2021.anzahl\_gerate\_2021 - U2013.anzahl\_gerate\_2013 AS zuwachs\_geraete}\\ 
& \texttt{FROM Untersuchungen2013 AS U2013}\\ 
& \texttt{JOIN Untersuchungen2021 AS U2021} \\ 
& \texttt{ON U2013.kanton = U2021.kanton2021}\\ 
& \texttt{ORDER BY zuwachs\_untersuchungen\_gesamt DESC}\\ 
& \texttt{LIMIT 3;}\\

\midrule

\colorbox{myYellowGreenBg}{\textcolor{myYellowGreenTx}{5. Numeric Operators mixed}} & \begin{tabular}[c]{@{}l@{}}{[}resident\_population\_birthplace\_citizenship\_type{]} \end{tabular} \\

\colorbox{myYellowGreenBg}{\textcolor{myYellowGreenTx}{with Aggregators}} \\

\addlinespace
& \begin{tabular}[c]{@{}p{12cm}@{}}\textit{What was the percentage of the population in Switzerland who were born in a foreign country on 2017?} \end{tabular} \\
\addlinespace
& \texttt{SELECT SUM(un.in), SUM(un.out),} \\
& \texttt{\textbf{\colorbox{myYellowGreenBg}{\textcolor{myYellowGreenTx}{SUM(un.out)/(SUM(un.in)+SUM(un.out))}}} AS percentage} \\ 
& \texttt{FROM (SELECT T1.year, T1.place\_of\_birth, T1.citizenship, }\\ 
& \texttt{SUM(T1.amount) AS in, SUM(0) AS out} \\ 
& \texttt{FROM resident\_population\_birthplace\_citizenship\_type AS T1}\\ 
& \texttt{JOIN spatial\_unit AS T2} \\ 
& \texttt{ON T1.spatialunit\_uid = T2.spatialunit\_uid}\\ 
& \texttt{WHERE T2.country = 'True' AND T1.year = 2017}  \\ 
& \texttt{AND T1.place\_of\_birth = 'Switzerland'} \\ 
& \texttt{AND T1.citizenship = 'Citizenship - total'}\\ 
& \texttt{GROUP BY T1.year,  T1.place\_of\_birth, T1.citizenship}\\ 
& \texttt{UNION}\\ 
& \texttt{SELECT T1.year,T1.place\_of\_birth,T1.citizenship,} \\ 
& \texttt{SUM(0) AS in, SUM(T1.amount) AS out} \\ 
& \texttt{FROM resident\_population\_birthplace\_citizenship\_type AS T1}\\ 
& \texttt{JOIN spatial\_unit AS T2} \\ 
& \texttt{ON T1.spatialunit\_uid = T2.spatialunit\_uid}\\ 
& \texttt{WHERE T2.country = 'True' AND T1.year = 2017}  \\ 
& \texttt{AND T1.place\_of\_birth = 'Abroad'} \\ 
& \texttt{AND T1.citizenship = 'Citizenship - total'}\\ 
& \texttt{GROUP BY T1.year, T1.place\_of\_birth, T1.citizenship)} \\  
& \texttt{AS un;} \\ 
\midrule

\colorbox{myRoyalBlueBg}{\textcolor{myRoyalBlueTx}{6. SET-Operation Query}} & \begin{tabular}[c]{@{}l@{}}{[}marriage\_citizenship{]}\end{tabular} \\
\addlinespace
& \begin{tabular}[c]{@{}p{12cm}@{}}\textit{In 2000, among the municipalities, which had the highest and lowest numbers of marriages where the husbands were Swiss citizens?} \end{tabular} \\
\addlinespace
& \texttt{(SELECT T2.name, T2.spatialunit\_ontology, T1.year, T1.amount  }\\ 
& \texttt{FROM marriage\_citizenship as T1}\\ 
& \texttt{JOIN spatial\_unit AS T2}\\ 
& \texttt{ON T1.spatialunit\_uid = T2.spatialunit\_uid}\\ 
& \texttt{WHERE T2.municipal = 'True'}\\ 
& \texttt{AND T1.year = 2000}\\ 
& \texttt{AND T1.citizenship\_category\_husband = 'Switzerland'}\\ 
& \texttt{AND T1.citizenship\_category\_wife = 'Citizenship of wife - total'}\\ 
& \texttt{AND T1.amount = (SELECT Max(T1.amount)} \\
& \texttt{FROM marriage\_citizenship as T1}\\ 
& \texttt{JOIN spatial\_unit AS T2} \\ 
& \texttt{ON T1.spatialunit\_uid = T2.spatialunit\_uid}\\ 
& \texttt{WHERE T2.municipal = 'True'} \\ 
& \texttt{AND T1.year = 2000} \\ 
& \texttt{AND T1.citizenship\_category\_husband = 'Switzerland'} \\ 
& \texttt{AND T1.citizenship\_category\_wife = 'Citizenship of wife - total')}\\ 
& \texttt{LIMIT 1)} \\ 
& \texttt{\textbf{\colorbox{myRoyalBlueBg}{\textcolor{myRoyalBlueTx}{UNION ALL}}}}\\ 
& \texttt{(SELECT T2.name, T2.spatialunit\_ontology, T1.year, T1.amount}\\ 
& \texttt{FROM marriage\_citizenship as T1}\\ 
& \texttt{JOIN spatial\_unit AS T2} \\ 
& \texttt{ON T1.spatialunit\_uid = T2.spatialunit\_uid}\\ 
& \texttt{WHERE T2.municipal = 'True'} \\ 
& \texttt{AND T1.year = 2000} \\ 
& \texttt{AND T1.citizenship\_category\_husband = 'Switzerland'} \\ 
& \texttt{AND T1.citizenship\_category\_wife = 'Citizenship of wife - total'}\\ 
& \texttt{AND T1.amount = (SELECT MIN(T1.amount)} \\ 
& \texttt{FROM marriage\_citizenship as T1}\\ 
& \texttt{JOIN spatial\_unit AS T2} \\ 
& \texttt{ON T1.spatialunit\_uid = T2.spatialunit\_uid}\\ 
& \texttt{WHERE T2.municipal = 'True'} \\ 
& \texttt{AND T1.year = 2000}  \\ 
& \texttt{AND T1.citizenship\_category\_husband = 'Switzerland'} \\ 
& \texttt{AND T1.citizenship\_category\_wife = 'Citizenship of wife - total')}\\ 
& \texttt{LIMIT 1);} \\ 
\midrule
\colorbox{myVioletBg}{\textcolor{myVioletTx}{7. Special Keywords}} & \begin{tabular}[c]{@{}l@{}}{[}stock\_vehicles{]}\end{tabular} \\
\addlinespace
& \begin{tabular}[c]{@{}p{12cm}@{}}\textit{What was the proportion of electric vehicles in Geneva in 2010?} \end{tabular} \\
\addlinespace
& \texttt{SELECT }\\
& \texttt{SUM(\colorbox{myVioletBg}{\textcolor{myVioletTx}{\textbf{CASE WHEN}}} sv.fuel\_type \colorbox{myVioletBg}{\textcolor{myVioletTx}{\textbf{ILIKE}}} '\%electric\%' \colorbox{myVioletBg}{\textcolor{myVioletTx}{\textbf{THEN}}} sv.amount \colorbox{myVioletBg}{\textcolor{myVioletTx}{\textbf{ELSE}}} 0 \colorbox{myVioletBg}{\textcolor{myVioletTx}{\textbf{END}}}) AS electric\_vehicles,}\\
& \texttt{SUM(sv.amount) AS total\_vehicles,}\\     
& \texttt{(SUM(CASE WHEN sv.fuel\_type \colorbox{myVioletBg}{\textcolor{myVioletTx}{\textbf{ILIKE}}} '\%electric\%'} \\
& \texttt{\colorbox{myVioletBg}{\textcolor{myVioletTx}{\textbf{THEN}}} sv.amount \colorbox{myVioletBg}{\textcolor{myVioletTx}{\textbf{ELSE}}} 0 \colorbox{myVioletBg}{\textcolor{myVioletTx}{\textbf{END}}}) * 100.0 / NULLIF(SUM(sv.amount), 0))} \\ 
& \texttt{AS proportion\_electric\_vehicles}\\ 
& \texttt{FROM spatial\_unit AS su}\\ 
& \texttt{INNER JOIN stock\_vehicles AS sv}\\ 
& \texttt{ON su.spatialunit\_uid = sv.spatialunit\_uid}\\ 
& \texttt{WHERE su.name \colorbox{myVioletBg}{\textcolor{myVioletTx}{\textbf{ILIKE}}} '\%geneva\%'}\\     
& \texttt{AND sv.year = 2010;}\\

\midrule
\colorbox{myRoseBg}{\textcolor{myRoseTx}{8. \texttt{NULL}-value}} & \begin{tabular}[c]{@{}l@{}}{[}nationalratswahlen{]}\end{tabular} \\
\addlinespace
& \begin{tabular}[c]{@{}p{12cm}@{}}\textit{In welchen Kantonen stand die Partei SVP 2019 nicht zur Wahl?} \end{tabular} \\
\addlinespace
& \texttt{SELECT S.name\_de AS kanton\_ohne\_svp\_2019} \\ 
& \texttt{FROM nationalratswahlen AS T}\\ 
& \texttt{JOIN spatial\_unit AS S} \\
& \texttt{ON T.spatialunit\_uid = S.spatialunit\_uid}\\ 
& \texttt{WHERE S.canton = 'TRUE'}\\   
& \texttt{AND T.partei = 'SVP'}\\   
& \texttt{AND T.jahr = 2019}\\   
& \texttt{AND T.parteistarke\_in\_prozent IS \colorbox{myRoseBg}{\textcolor{myRoseTx}{\textbf{NULL}}}}; \\
\bottomrule
\end{longtable}
}
\newpage

\section{Failure Case}\label{app:C}
Upon examining the failed prediction of easy queries, we identified the root cause as a key term swap between \texttt{COUNT} and \texttt{DISTINCT} during the few-shot ICL execution on Mixtral. As shown below, the discrepancy between the expected and predicted queries is evident for the Mixtral setting at an easy difficulty level in the few-shot scenario.
\begin{table}[htbp]
\small
  \centering
 
  \label{tab:data}
  \begin{tabular}{@{}ll@{}}
    \toprule
    \textbf{Data Schema}        & baby\_names\_favorite\_firstname \\
    \midrule
    \textbf{Question}           & How many favorite baby names are registered in year 2011? \\
    \midrule
    \textbf{Ground Truth}       & \texttt{SELECT COUNT(DISTINCT first\_name) FROM baby\_names\_favorite\_firstname} \\
            & \texttt{WHERE year = 2011} \\
    \midrule

    \textbf{Predicted Query}    & \texttt{SELECT DISTINCT COUNT(bnff.first\_name) FROM baby\_names\_favorite\_firstname } \\
        &\texttt{as bnff WHERE bnff.year = 2011} \\
  
    \bottomrule
  \end{tabular}
\end{table}

\section{Detailed Results}\label{app:E}
\begin{table*}[htbp]
\setlength\tabcolsep{4pt}
  \centering

  \begin{tabular}{p{2cm}|c|ccc|ccc}
    \toprule
    \multirow{2}{*}{Experiments}& \multirow{2}{*}{Shot}& \multicolumn{3}{c|}{GPT-3.5} & \multicolumn{3}{c}{Mixtral} \\
    \cmidrule(lr){3-5} \cmidrule(lr){6-8} 
  & & {$\text{EA}_\text{strict}$} & {$\text{EA}_\text{soft}$} &{$\text{EA}_\text{partial}$}& {$\text{EA}_\text{strict}$} & {$\text{EA}_\text{soft}$} &{$\text{EA}_\text{partial}$} \\
  
    \midrule
    Zero-shot & 0 & 13.52 (0.33) & 13.76 (0.33) & 17.95 (0.33) & 5.82 (0.33)  &  5.82 (0.33) & 6.52 (0.33) \\
     \cmidrule(lr){1-8}
 
  & 1 & 36.36 (0.00) & 41.26 (0.00) & 46.15 (0.00)  & 18.18 (0.00) & 21.68 (0.00) & 23.08 (0.00) \\
    & 2 & 36.36 (0.44) & 41.68 (0.34) & 46.57 (0.34)  & 21.68 (0.57) & 27.97 (0.57) & 29.37 (0.57) \\
  \multirow{2}{*}{Few-shot} & 3 & 35.02 (0.81) & 41.12 (0.52) & 46.01 (0.52)  & 23.78 (0.44) & 29.23 (1.03) & 31.75 (1.75) \\
     \multirow{2}{*}{\scriptsize{(Random)}}& 4 & 36.36 (0.44) & 41.26 (0.44) & 43.50 (0.52) & \textbf{24.20} (0.56) & \textbf{30.07} (0.89) & \textbf{32.87} (0.89) \\
  & 5 & 36.83 (0.33) & \textbf{44.53}(0.33) & \textbf{50.58} (0.66) & {23.78} (0.57) & \textbf{30.07} (0.57) & 32.40 (0.87) \\
   & 6 & \textbf{37.76} (0.00) & 42.66 (0.00) & 45.45 (0.00)& 21.68 (0.00) & 26.57 (0.00) & 29.84 (0.33) \\
   & 8 & 36.36 (0.00) & 41.26 (0.00) & 44.76 (0.00)  &  21.21 (0.33) & 29.60 (0.33) & 31.94 (0.33) \\
    \cmidrule(lr){1-8}
     &1 & 33.57 (0.00) & 38.46 (0.00) & 41.26 (0.00) & 16.92 (0.28) & 23.78 (0.00) & 26.57 (0.00) \\
   & 2 & 33.71 (0.28) & 38.60 (0.28) & 40.56 (0.00)& 21.12 (1.12) & 26.71 (1.12) & 28.95 (0.95) \\
  \multirow{2}{*}{Few-shot} & 3 & 39.02 (0.28) & 45.73 (0.56) & 47.83 (0.56) & 22.52 (0.82) & 28.81 (0.82) & 30.63 (0.69) \\
  \multirow{2}{*}{\scriptsize{(Similarity)}} & 4 & 39.16 (0.44) & 46.15  (0.44) & 48.25 (0.44)& 26.01 (0.69) & 32.31 (0.69) & 34.13 (0.82) \\
   & 5 & \textbf{41.68}  (0.56) & \textbf{48.25} (0.44) & \textbf{50.07} (0.34) & 21.68 (0.44) & 30.77 (0.44) & 35.94 (0.71) \\
   & 6 & 40.56 (0.44)& 45.45 (0.44) & 46.85 (0.44)& \textbf{28.39} (0.34) & \textbf{35.38} (0.34) & \textbf{38.18} (0.34) \\
   & 8 & 39.30 (0.52) & 44.90 (0.52) & 46.99 (0.52)& 21.40 (0.34) & 28.95 (0.34) & 31.47 (0.44) \\
   
    \bottomrule
  \end{tabular}
  \caption{Execution accuracy for different query evaluation metrics (strict, soft and partial). The upper part shows few-shot results where the samples are \emph{randomly} selected. The bottom part shows few-shot results where the samples are selected based on a \emph{similarity score}. Accuracy is presented by the average and standard deviation across three separate and independent runs.}
   \label{tab:table-result-shots}
\end{table*}

\begin{table*}[htbp]
  \centering
  
  \begin{tabular}{l|l|cc|cc}
    \toprule
  \multirow{2}{*}{Experiment}& \multirow{2}{*}{Hadness}& \multicolumn{2}{c|}{GPT-3.5} & \multicolumn{2}{c}{Mixtral} \\
\cmidrule(lr){3-4} \cmidrule(lr){5-6} 
    &  & EN & DE & EN & DE \\
 \midrule
  \multirow{5}{*}{Zero-shot}&Easy \texttt{(2,0)} & 100.00 (0.00)  &0.00 (0.00)  & 100.00 (0.00)  & 0.00 (0.00)  \\
  &Medium \texttt{(8,8)} & 29.17 (5.89) & 37.50 (0.00) & 41.67 (5.89) & 12.50 (0.00)  \\
  &Hard \texttt{(9,9)} & 11.11 (0.00)  & 22.22 (0.00)  & 0.00 (0.00)  & 11.11 (0.00)   \\
  &Extra hard \texttt{(24,24)}& 0.00 (0.00)  & 29.17 (0.00)   & 4.17 (0.00)  &  0.00 (0.00)  \\
  &Unknown \texttt{(18,41)}& 0.00 (0.00)  & 4.88 (0.00)  &  0.00 (0.00)  &  0.00 (0.00)  \\
  \midrule
\multirow{2}{*}{Few-shot}&Easy \texttt{(2,0)} &  100.0 (0.00) & 0.00 (0.00) & 50.00 (0.00) & 0.00 (0.00)  \\
 \multirow{2}{*}{\scriptsize{}{(similarity)}} &Medium \texttt{(8,8)}&  75.00 (0.00) & 50.00 (0.00) & 45.00 (6.12) & 37.50 (0.00)   \\
  &Hard \texttt{(9,9)}& 11.11 (0.00) & 44.44 (0.00) &   11.11 (0.00) & 33.33 (0.00)   \\
  &Extra hard \texttt{(24,24)} &  29.17 (2.50)& 50.00 (0.00)  &  25.00 (0.00) & 45.83 (0.00)  \\
  &Unknown \texttt{(18,41)} & 27.78 (0.00) & 44.88 (1.20)& 11.11 (0.00) & 24.39 (0.00)  \\
  \bottomrule
  \end{tabular}
  \caption{Strict execution accuracy ($\text{EA}_\text{strict}$) of zero-shot and optimal few-shot results per query hardness (easy, medium, hard, extra hard, and unknown).}
  \label{tab:table-result-hardness}
\end{table*}

\begin{table*}[t]
  \centering
 
  \begin{tabular}{@{}l|l|cc|cc@{}}
    \toprule
   \multirow{2}{*}{Experiment} & \multirow{2}{*}{Execution Accuracy (EA)}& \multicolumn{2}{c|}{GPT-3.5} & \multicolumn{2}{c}{Mixtral} \\
   \cmidrule(lr){3-4} \cmidrule(lr){5-6}
    && EN \texttt{(\#82)}  & DE \texttt{(\#61)} & EN \texttt{(\#82)} & DE \texttt{(\#61)} \\ \midrule
    \multirow{3}{*}{Zero-shot}& strict & 8.75 (0.77) & 17.07 (0.00) & 10.39 (0.77) & 2.44 (0.00) \\
    &soft & 8.75 (0.77) & 17.48 (0.58) & 10.39 (0.77) & 2.44 (0.00) \\
      &partial & 10.39 (0.77) & 23.17 (0.00) &12.02 (0.77) & 2.44 (0.00) \\
      \midrule
     \multirow{3}{*}{Few-shot} &strict& 34.43 (0.00) & 46.83 (0.60) & 22.29 (0.80) & 32.93 (0.00) \\
     &soft & 40.00 (0.80) & 54.15 (0.60) & 27.21 (0.80) & 41.46 (0.00) \\ 
      &partial& 41.67 (0.66) & 57.08 (0.49) & 30.49 (0.80) & 43.90 (0.00) \\\bottomrule
  \end{tabular}
   \caption{Mean execution accuracy (standard deviation) using three different metrics (strict, soft and partial) for zero-shot and optimal few-shot experiments. The results are shown for the two different languages English 
   (EN) and German (DE). \texttt{\#XX} represents the number of NL/SQL-pairs in the development set for each language.}
  \label{tab:table-result-best-few-shots}
\end{table*}

\begin{figure*}[bhp]
    \includegraphics[width=1\textwidth]{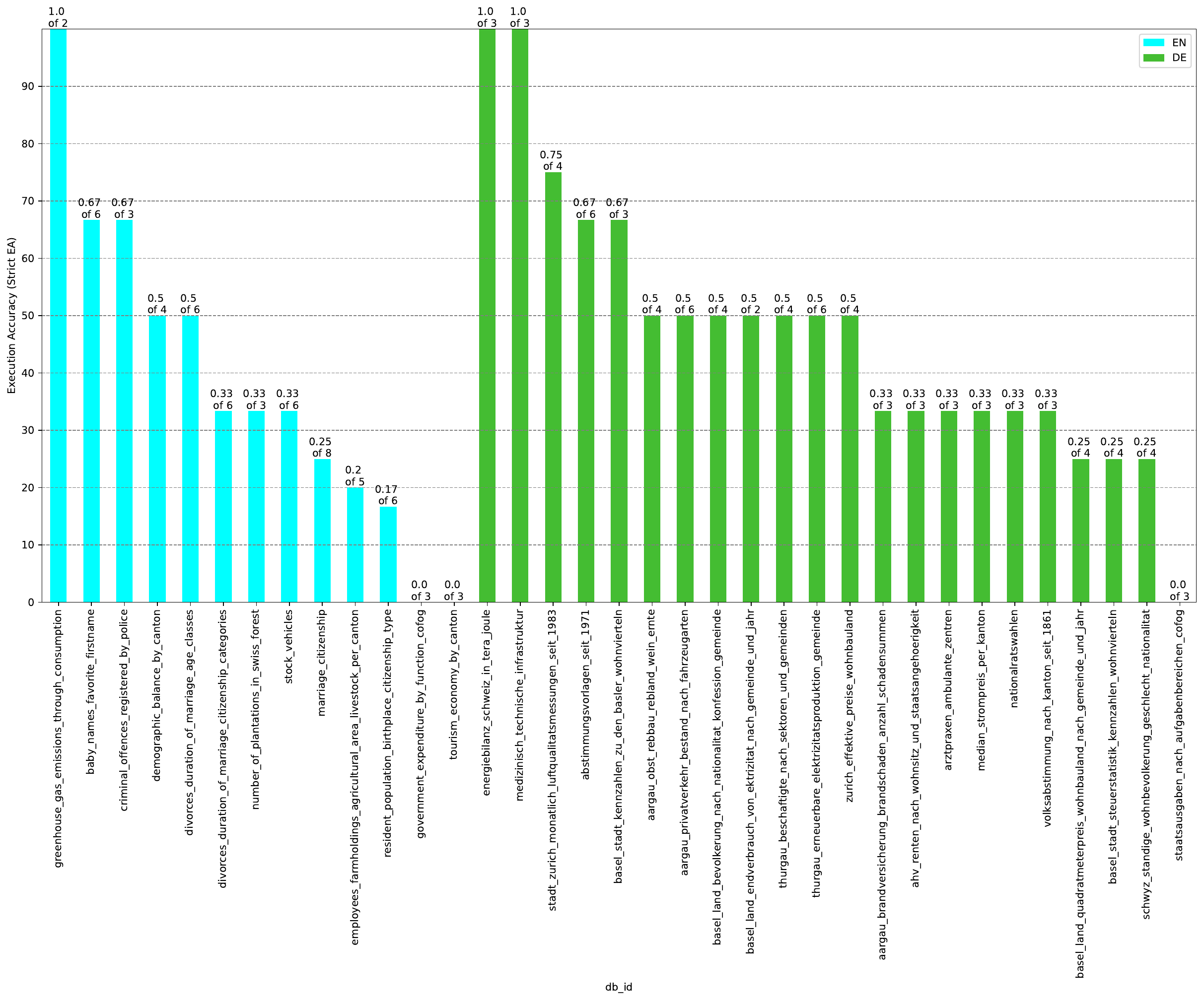}
    \caption{Strict execution accuracy  ($\text{EA}_\text{strict}$) per knowledge domain and language over 35 different databases. Left hand: English. Right hand: German.}
    \label{fig:HEA_dataset}
\end{figure*}

\end{document}